\documentclass[pmlr]{jmlr}

\usepackage{longtable}

\usepackage{booktabs}
\usepackage[load-configurations=version-1]{siunitx} 

\makeatletter
\def\set@curr@file#1{\def\@curr@file{#1}} 
\makeatother

\theorembodyfont{\upshape}
\theoremheaderfont{\scshape}
\theorempostheader{:}
\theoremsep{\newline}

\usepackage{enumitem}
\usepackage{algorithm}
\usepackage{graphics}
\usepackage{algorithmic}
\usepackage{multirow}
\usepackage{adjustbox}

\title[]{Learning by Ignoring, with Application to Domain Adaptation}

\author{\Name{Xingchen Zhao\textsuperscript{\dag}
}
       \Email{XIZ168@pitt.edu} 
       \AND
       \Name{Xuehai He\textsuperscript{\dag}
}
       \Email{x5he@eng.ucsd.edu } 
       \AND
       \Name{Pengtao Xie\textsuperscript{*}}
       \Email{p1xie@eng.ucsd.edu}\\
       \addr 
University of California San Diego
\AND
       }

\begin{document}

\maketitle

\begin{abstract}
Learning by ignoring, which identifies less important things and excludes them from the learning process, is broadly practiced in human learning and has shown ubiquitous effectiveness. 
There has been psychological studies showing that  learning to ignore certain things is a powerful tool for helping people focus. In this paper, we explore  whether this useful human  learning methodology can be borrowed  to improve machine learning. We propose a novel machine learning framework referred to as learning by ignoring (LBI). Our framework automatically identifies pretraining data examples that have large domain shift from the target distribution by learning an ignoring variable for each example and excludes them from the pretraining process.  We formulate LBI as a three-level optimization framework  where  three learning stages are involved: pretraining by minimizing the losses weighed by ignoring variables; finetuning; updating the ignoring variables by minimizing the validation loss.   An gradient-based algorithm is developed to efficiently solve the three-level optimization problem in LBI.  Experiments on various datasets demonstrate the effectiveness of our framework.
\end{abstract}

\section{Introduction}

\let\thefootnote\relax\footnotetext{$^\dag$Equal contribution.}
\let\thefootnote\relax\footnotetext{$^*$Corresponding author.}

In human learning, a widely-practiced effective learning methodology is learning by ignoring. 
For example,  in course learning, given a large collection of practice problems provided in the textbook, the teacher selects a subset of problems as homework for the students to practice instead of using all problems in the textbook. Some practice problems are ignored because 1) they are too difficult which might confuse the students; 2) they are too simple which are not effective in helping the students to practice their knowledge learned during lectures; 3) they are repetitive. The study in~\citep{cunningham2016taming} shows that learning to ignore certain things is powerful for helping people focus.

\begin{figure}[t]
    \centering
 \includegraphics[width=0.8\columnwidth]{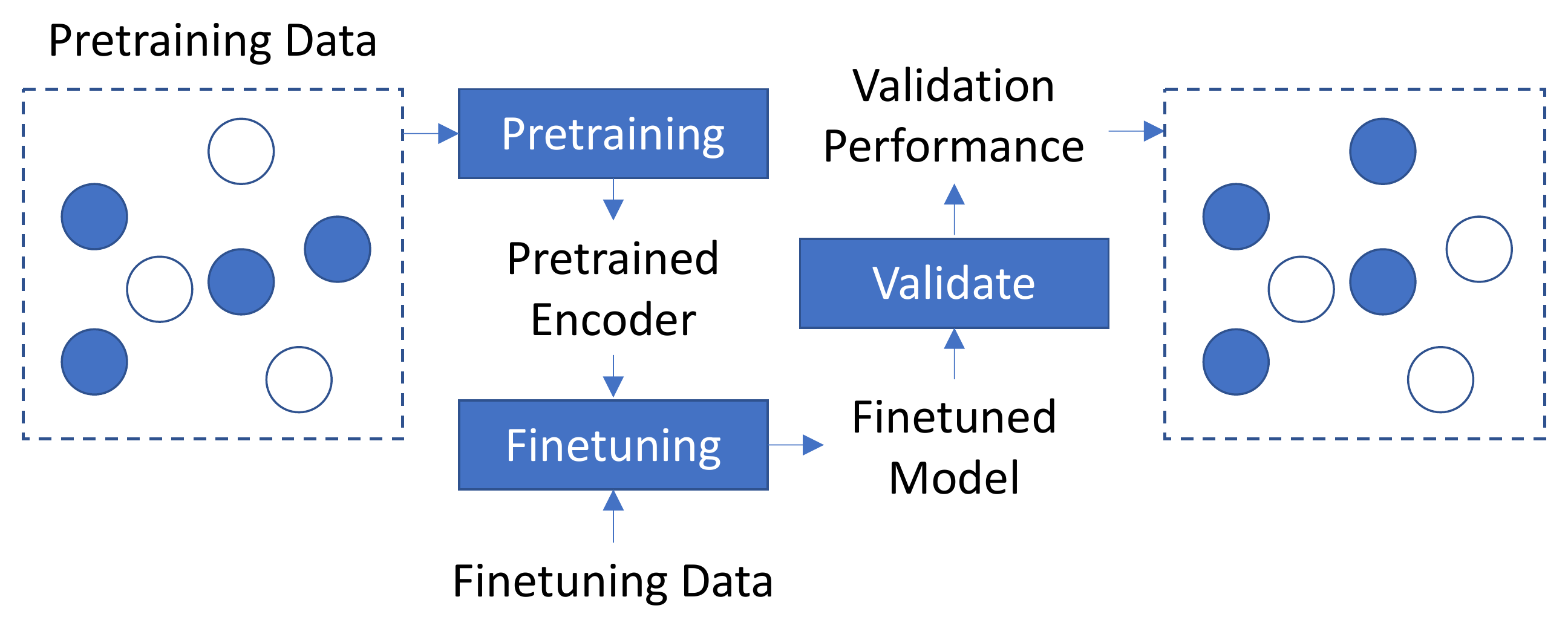}
       \caption{Illustration of learning by ignoring. Void circles denote ignored pretraining data examples. Given a set of intermediately  selected pretraining examples, they are used to pretrain a model, which is then finetuned and validated. The validation performance provides guidance on what pretraining examples should be ignored in the next round of learning. This process iterates until convergence. }
 \label{fig:illus}
\end{figure}

Drawing inspirations from this effective human-learning method,  
we are intrigued in exploring whether this method is helpful for training better machine learning models as well. We propose a novel machine learning framework called learning by ignoring (LBI) (as illustrated in Figure~\ref{fig:illus}). In this framework, a model is trained to perform a target task. The model consists of a data encoder and a task-specific head. The encoder is trained in two phrases: pretraining  and finetuning, conducted on a pretraining dataset and a finetuning dataset. Pretraining is a commonly used technique in deep learning to learn more effective representations for alleviating overfitting. Given a target task where the amount of training data is limited, training deep neural networks on this small-sized dataset has high risk of overfitting. To address this problem, one can pretrain the feature extraction layers in the network on large-sized external data from some source domain, then finetune these layers on the target data. The abundance of source data enables the network to learn powerful representations that are robust to overfitting. And such representation power can be leveraged to assist in the learning of the target task with more resilience to overfitting.

Some pretraining data examples have a domain difference with the finetuning dataset, rendering them not suitable to pretrain the data encoder that will be used for performing the target task. We would like to identify such out-of-domain data examples and exclude them from the pretraining process. To achieve this goal, we associate each pretraining example with an ignoring variable $a\in[0,1]$. If $a$ is close to 0, it means that this example should be ignored. We develop a three-level framework (involving three learning stages) to automatically learn these ignoring variables. In the first learning stage, we pretrain a data encoder $V$ by minimizing a weighted pretraining loss: the loss defined on each pretraining example is multiplied with the ignoring variable of this example. If an example $x$ should be ignored, its $a$ is close to 0; multiplying $a$ to the loss of $x$ makes this loss close to 0, which effectively excludes $x$ from the pretraining process. In this stage, these ignoring variables $A=\{a\}$ are fixed. They will be updated at a later stage. Note that the optimally pretrained encoder $V^*(A)$ is a function of $A$ since $V^*(A)$ is a function of the weighted loss and the weighted loss is a function of $A$. In the second stage, we train another data encoder $W$ on the finetuning dataset. During the training of $W$, we encourage it to be close to the optimal encoder $V^*(A)$ trained in the first stage by minimizing the squared L2 distance between $W$ and $V^*(A)$. $V^*(A)$ contains information distilled from non-ignored pretraining examples. Encouraging $W$ to be close to $V^*(A)$ effectively achieves the goal of pretraining $W$ on these non-ignored examples. Note that the optimally trained encoder $W^*(V^*(A))$ is a function of $V^*(A)$. In the third stage, we apply $W^*(V^*(A))$ to make predictions on the validation set of the target task and update $A$ by minimizing the validation loss. To summarize, we aim to learn an ignoring variable for each pretraining example so that the data encoder pretrained on non-ignored examples achieves the optimal performance on the validation set after finetuning. The three stages are conducted end-to-end in a joint manner. 
Experiments on various datasets demonstrate the effectiveness of our method.

The major contributions of this paper is as follows:
\begin{itemize}
\item Drawing inspirations from the  ignoring-driven learning methodology of humans, we propose a novel machine learning framework called learning by ignoring (LBI). Our framework automatically identifies pretraining data examples that have large domain shift from the target distribution by learning an ignoring variable for each example and excludes them from the pretraining process. 
\item We formulate LBT as  a multi-level optimization framework  involving three learning stages: pretraining by minimizing the losses weighed by ignoring variables; finetuning; updating the ignoring variables by minimizing the validation loss. 
\item An efficient gradient-based algorithm is developed to solve the LBI problem. 
\item Experiments on various datasets demonstrate the effectiveness of our method. 
\end{itemize}

\section{Related Works}
\paragraph{Data Re-weighting and Selection.}
Several approaches have been proposed for data selection. Matrix column subset selection~\citep{deshpande2010efficient,boutsidis2009improved} aims to select a subset of data examples that can best reconstruct the entire dataset. Similarly, coreset selection~\citep{bachem2017practical} chooses representative training examples in a way that models trained on the selected examples have comparable performance with those trained on all training examples.  These methods perform data selection and model training separately.  As a result, the validation performance of the model cannot be used to guide data selection. \citet{ren2018learning} proposes a meta learning method to learn the weights of  training examples by performing a meta gradient descent step on the weights of the current mini-batch of examples. \citet{shu2019meta} propose a method which can adaptively learn an explicit weighting function
directly from data. These works focus on selecting training (finetuning) examples using a  bi-level optimization framework while our work focuses on selecting pretraining examples using a three-level optimization framework. 

\paragraph{Pretraining.}
Arguably, the most popular pretraining approach is supervised pretraining (SP)~\citep{tl}, which learns the weight parameters of a representation network by solving a supervised source task, i.e., correctly mapping input data examples to their labels (e.g., classes, segmentation masks, etc.).  Recently, self-supervised learning~\citep{oord2018representation,he2019momentum,chen2020simple,misra2020self}, as an unsupervised pretraining approach, has achieved promising success and outperforms supervised pretraining in a wide range of applications. Similar to SP, self-supervised pretraining (SSP) also solves predictive tasks. But the output labels in SSP are constructed from the input data, rather than annotated by humans as in SP. The auxiliary predictive tasks in SSP could be predicting whether two augmented data examples originate from the same original data example~\citep{henaff2019data,he2019momentum,chen2020simple,misra2020self}, inpainting masked regions in images~\citep{inpainting}, etc. 

\paragraph{Domain Adaptation.} Domain adaptation~\citep{da1,da2,da3} considers the problem
of transferring knowledge between two domains with distinctive data distributions. There are mainly two approaches to address the domain adaptation problem. One way is by using metric learning, where a distance metric is defined to measure the distribution discrepancy between domains and the target of training is to minimize the distance~\citep{md1,md2,md3}. Another way is by doing adversarial domain adaptation~\citep{ad1,ad2}. Instead of resorting to metric learning which explicitly optimizes
a similarity function, it learns a domain
discriminator and a feature learning network adversarially. The domain discriminator is trained to tell whether an instance is from the
source domain or the target domain, while the feature learning network learns domain-invariant features and is trained to fool the domain discriminator. 

Apart from the classic domain adaptation settings, in recent years, there are also works focusing on unsupervised domain adaptation (UDA), which gives predictions for the unlabeled target data, where labels only exist in the source data.  There are mainly three types of methods for unsupervised domain adaptation in computer vision. 
The first one uses a model trained on the labeled source data to estimate labels on
the target data, then trains the model using pseudo target labels~\citep{contrastive}; the second one is to induce alignment between the source and the target domains in feature spaces, e.g. ~\citep{mm1,mm2} propose to minimize the Maximum Mean Discrepancy (MMD) between
the source and target domain in the deep neural network; The third type uses generative models to transform the source images to resemble the target images~\citep{gan_uda}, which operates on image pixels rather than on an intermediate representation space like the first type of methods. Different from these approaches for UDA, in this paper, we use labels both from source dataset and target dataset. We are also the first to use learning by ignoring on domain adaptation problems.

\section{Methods}

Our framework aims to learn a machine learning model for accomplishing a target task $T$. The ML model is composed of 
a data encoder $W$ and a task-specific head $H$.
For instance, in a text classification task, the data encoder can be a BERT~\citep{devlin2018bert} model which produces an embedding of the input text and the classification head can be a multi-layer perceptron which predicts the class label of this text based on its embedding.  
The learning is performed in two phrases: pretraining and finetuning. In the pretraining phrase, we pretrain the data encoder $W$ on a pretraining dataset $D^{\textrm{(pre)}}=\{d^{\textrm{(pre)}}_i\}_{i=1}^M$ by solving a pretraining task $P$. $P$ could be the same as $T$. In this case, $W$ and the task-specific head $H$ are  trained jointly on $D^{\textrm{(pre)}}$. $P$ could be  different from $T$ as well. Under such circumstances, $P$ has its own task-specific head $J$ while sharing the same encoder $W$ with $T$. $J$ and $W$ are  trained jointly on $D^{\textrm{(pre)}}$. Oftentimes, some pretraining data examples have a domain shift with the finetuning dataset. This domain shift renders these examples not suitable for pretraining the encoder. We aim to automatically identify such examples using ignoring variables and exclude them from the pretraining process. To achieve this goal, we multiply the loss of a pretraining example $x$ with an ignoring variable $a\in[0,1]$. If $a$ is close to 0, it means that this example should be ignored; accordingly, the loss (after multiplied with $a$) is made close to 0, which effectively excludes $x$ from the pretraining process. We aim to automatically learn the values of these ignoring variables, which will be detailed later. After pretraining, the encoder is finetuned on the training dataset $D^{\textrm{(tr)}}=\{d^{\textrm{(tr)}}_i\}_{i=1}^N$. For mathematical convenience, we formulate ``pretraining'' in the following way: in the pretraining phrase, we train another encoder $V$; in the finetuning phrase, we encourage the encoder $W$ to be close to the optimally pretrained encoder $V^*$  where the closeness is measured using squared L2 distance $\|W-V^*\|_2^2$.

Overall,  the learning is performed in three stages. In the first stage, the model trains the encoder $V$ and the head $J$ specific to the pretraining task $P$ on the pretraining dataset $D^{\textrm{(pre)}}=\{d^{\textrm{(pre)}}_i\}_{i=1}^M$, with the ignoring variables $A=\{a_i\}_{i=1}^{M}$ fixed:
\begin{equation}
    V^*(A)=\textrm{min}_{V,J}\;\;\sum_{i=1}^{M} a_i L(V,J, d^{(\textrm{pre})}_i).
\end{equation}
After training, the optimal head is discarded. The optimal encoder $V^*(A)$ is retained for future use. 
The ignoring variables $A$ are needed to make predictions and calculate training losses.  
But they should not be updated in this stage. Otherwise, the values of $A$ will all be zero. Note that $V^*(A)$  is a function of $A$ since it is a function of $\sum_{i=1}^{M} a_i L(V,J, d^{(\textrm{pre})}_i)$ and $\sum_{i=1}^{M} a_i L(V,J, d^{(\textrm{pre})}_i)$ is a function of $A$.

\begin{table}[t]
\caption{Notations in learning by ignoring}
\centering
\begin{tabular}{l|l}
\hline
Notation & Meaning \\
\hline
$M$ & Number of pretraining examples\\
$N$ & Number of finetuning examples\\
$O$ & Number of validation examples\\
$d_i^\textrm{(pre)}$ & The $i$-th pretraining data example\\
$a_i$ & Ignoring variable of $d_i^\textrm{(pre)}$ in pretraining\\
$b_i$ & Ignoring variable of $d_i^\textrm{(pre)}$ in finetuning\\
$d_i^\textrm{(tr)}$ & The $i$-th  finetuning data example \\
$d_i^\textrm{(val)}$ & The $i$-th  validation example \\
$W$ & Encoder in finetuning\\
$V$ & Encoder in pretraining\\
$H$ & Head of the target task\\
$J$ & Head of the pretraining task\\
\hline
\end{tabular}
\label{tb:notations}
\end{table}

In the second stage, the model trains its data encoder $W$ and task-specific head $H$ by minimizing the training loss of the target task $T$. During training, $W$ is encouraged to be close to $V^*(A)$ trained in the pretraining phrase by minimizing the squared L2 distance $\|W-V^*(A)\|_2^2$. This implicitly achieves the effect of pretraining $W$ on the non-ignored pretraining examples. The optimization problem in the second stage is: 
\begin{equation}
\begin{array}{l}
    W^*(V^*(A)), H^*=
    \underset{W,H}{\textrm{min}}
    \;\;\sum\limits_{i=1}^{N}  L(W,H, d^{(\textrm{tr})}_i)+\lambda\|W-V^*(A)\|_2^2,
    \end{array}
    \label{eq:s2}
\end{equation}
where $\lambda$ is a tradeoff parameter.
Note that $W^*(V^*(A))$ is a  function of $V^*(A)$ since it is a function of $\|W-V^*(A)\|_2^2$, which is a function of $V^*(A)$.

In the third stage, we use the trained model consisting of $W^*(V^*(A))$ and $H^*$ to make predictions on the validation dataset $D^{(\textrm{val})}$ of the target task $T$. We update $A$ by minimizing the validation loss. 
\begin{equation}
    \textrm{min}_{A}\;\;\sum_{i=1}^{O}  L(W^*(V^*(A)), H^*, d^{(\textrm{val})}_i).
\end{equation}
The three stages mutually influence each other: $V^*(A)$ trained in the first stage is needed to calculate the loss function in the second stage; $W^*(V^*(A))$  trained in the second stage is needed to calculate the objective function in the third stage;  the ignoring variables $A$ updated in the third stage alter the loss function in the first  stage, which subsequently changes $V^*(A)$ and $W^*(V^*(A))$ as well.

\begin{figure}[t]
    \centering
 \includegraphics[width=0.7\columnwidth]{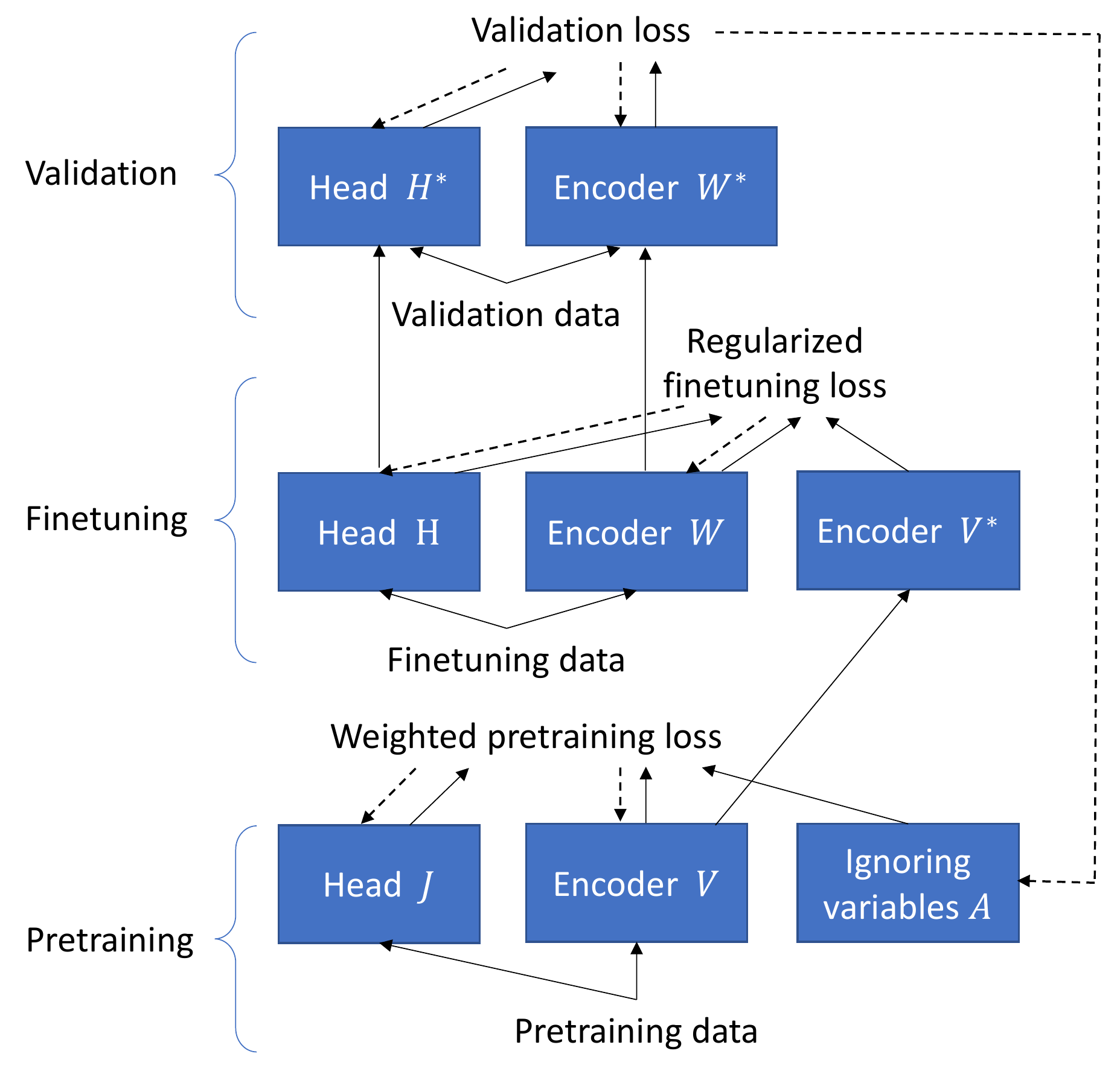}
       \caption{Learning by ignoring. Along the solid arrows, predictions are made and train/validation losses are calculated. Along the dotted arrows, gradients are calculated and parameters (including ignoring variables and network weights) are updated. 
       }
 \label{fig:arch}
\end{figure}

Putting the three learning stages together, we formulate LBI as the following three-level optimization problem:
\begin{equation}
    \begin{array}{l}
 \textrm{min}_{A}\;\;\sum_{i=1}^{O}  L(W^*(V^*(A)), H^*, d^{(\textrm{val})}_i)\\
s.t. \quad\;\;        W^*(V^*(A)), H^*=
    \textrm{min}_{W,H}\;\;\sum_{i=1}^{N}  L(W,H, d^{(\textrm{tr})}_i)+\lambda\|W-V^*(A)\|_2^2\\
\quad\quad\;\;\;     V^*(A)=\textrm{min}_{V,J}\;\;\sum_{i=1}^{M} a_i L(V,J, d^{(\textrm{pre})}_i)
\end{array}
\label{eq:lbi}
\end{equation}
This formulation consists of three optimization problems. The two inner optimization problems (on the constraints) represent the first and second learning stage respectively. The outer optimization problem represents the third learning stage. The three stages are illustrated in Figure~\ref{fig:arch}.

If the pretaining task and target task are the same, in the second stage we can use the pretraining data to train $W$ as well. Due to domain difference, not all pretraining examples are suitable for training $W$. To exclude such examples, we associate each pretraining example $d^{(\textrm{pre})}_i$ with an ignoring variable $b_i\in[0,1]$. Note that $b_i$ is different from $a_i$. $b_i$ is used to determine whether $d^{(\textrm{pre})}_i$ should be ignored during training $W$ and $a_i$ is used to determine whether $d^{(\textrm{pre})}_i$ should be ignored during training $V$. 
The corresponding formulation is:
\begin{equation}
    \begin{array}{l}
    \underset{A,B}{\textrm{min}}
 \;\;\;\;\sum_{i=1}^{O}  L(W^*(V^*(A),B), H^*(B), d^{(\textrm{val})}_i)\\
s.t. \quad\;\;        W^*(V^*(A),B), H^*(B)=
\underset{W,H}{\textrm{min}}
    \;\;\sum\limits_{i=1}^{N}  L(W,H, d^{(\textrm{tr})}_i)+ \lambda\|W-V^*(A)\|_2^2+
     \gamma \sum\limits_{i=1}^{M} b_i L(W,H, d^{(\textrm{pre})}_i)\\
\quad\quad\;\;\;     V^*(A)=
\underset{V,J}{\textrm{min}}\;\;\sum_{i=1}^{M} a_i L(V,J, d^{(\textrm{pre})}_i)
\end{array}
\label{eq:lbi-2}
\end{equation}
where $\gamma$ is a tradeoff parameter and $B=\{b_i\}_{i=1}^{M}$. Note that $W^*$ and $H^*$ are both functions of $B$. 

\subsection{Optimization Algorithm}
In this section, we develop a gradient-based optimization algorithm to solve the three-level optimization problem in Eq.(\ref{eq:lbi}). Drawing inspirations from~\citep{liu2018darts}, we approximate $V^{*}(A)$ using:
\begin{equation}
   V'= V-\xi_{V} \nabla_{V}\sum_{i=1}^{M} a_i L(V,J, d^{(\textrm{pre})}_i), 
   \label{eq:update_v}
\end{equation}
where $\xi_{V}$ is a learning rate. We update $J$ as:
\begin{equation}
    J\gets J-\xi_{J} \sum_{i=1}^{M} a_i \nabla_{J}L(V,J, d^{(\textrm{pre})}_i).
    \label{eq:update_j}
\end{equation}
Substituting  $V'$ into $\sum_{i=1}^{N}  L(W,H, d^{(\textrm{tr})}_i)+\lambda\|W-V^*(A)\|_2^2$, we obtain an approximated objective $\sum_{i=1}^{N}  L(W,H, d^{(\textrm{tr})}_i)+\lambda\|W-V'\|_2^2$. Then  we calculate the gradient of $O_{W}$ w.r.t to $W$ and approximate  $W^*(V^*(A))$ using one-step gradient descent update of  $W$:
\begin{equation}
\begin{array}{lll}
    W' &=& W-\xi_{W} \nabla_{W} (\sum_{i=1}^{N}  L(W,H, d^{(\textrm{tr})}_i)+ \lambda\|W-V'\|_2^2 )\\
    \end{array}
    \label{eq:update_w}
\end{equation}
Similarly, we approximate $H^*$ with:
\begin{equation}
\begin{array}{lll}
H'&=&H-\xi_{H} \nabla_{H} (\sum_{i=1}^{N}  L(W,H, d^{(\textrm{tr})}_i)+ \lambda\|W-V'\|_2^2 )\\
    \end{array}
    \label{eq:update_h}
\end{equation}
Finally, we substitute   $W'$ and $H'$ into $ \sum_{i=1}^{O}  L(W^*(V^*(A)), H^*, d^{(\textrm{val})}_i)$ and get $O_A=\sum_{i=1}^{O} $\\$ L(W', H', d^{(\textrm{val})}_i)$. We calculate the gradient of $O_A$ w.r.t $A$ and update $A$ by descending the gradient:
\begin{equation}
\begin{array}{lll}
    A &\gets &  A- \xi_{A}\nabla_{A}  \sum_{i=1}^{O}L(W',H', d^{(\textrm{val})}_i)\\
    \label{eq:update_a}
    \end{array}
\end{equation}
where
\begin{equation}
\begin{array}{lll}
    \frac{\partial V'}{\partial A}&=&\frac{\partial (V-\xi_{V} \nabla_{V}\sum_{i=1}^{M} a_i L(V,J, d^{(\textrm{pre})}_i) )}{\partial A} \\
    &=&-\xi_{V} \nabla^2_{A,V}\sum_{i=1}^{M} a_i L(V,J, d^{(\textrm{pre})}_i),
    \end{array}
\end{equation}
and
\begin{equation}
\begin{array}{lll}
    \frac{\partial W'}{\partial V'} & =&\frac{\partial (W-\xi_{W} ( \sum_{i=1}^{N}  \nabla_{W}L(W,H, d^{(\textrm{tr})}_i)+ 2\lambda(W-V') ))}{\partial V'}\\
    &=&2\xi_{W}\lambda I,
    \end{array}
\end{equation}
where $I$ is an identity matrix.

For the formulation in Eq.(\ref{eq:lbi-2}), the optimization algorithm is similar. $V^*(A)$ is approximated using Eq.(\ref{eq:update_v}). We approximate $W^*(V^*(A),B)$ using
\begin{equation}
\begin{array}{r}
    W'=W-\xi_{W} \nabla_{W} (\sum\limits_{i=1}^{N}  L(W,H, d^{(\textrm{tr})}_i)+\lambda\|W-V^*(A)\|_2^2+ \gamma \sum\limits_{i=1}^{M} b_i L(W,H, d^{(\textrm{pre})}_i)),
    \end{array}
\end{equation}
and approximate $H^*(B)$ using
\begin{equation}
\begin{array}{r}
    H'=H-\xi_{H} \nabla_{H} (\sum\limits_{i=1}^{N}  L(W,H, d^{(\textrm{tr})}_i)+\lambda\|W-V^*(A)\|_2^2+ \gamma \sum\limits_{i=1}^{M} b_i L(W,H, d^{(\textrm{pre})}_i)).
    \end{array}
\end{equation}
The update of $A$ is the same as that in Eq.(\ref{eq:update_a}). The update of $B$ is as follows:
\begin{equation}
\begin{array}{lll}
    B &\gets &  B- \xi_{B}\nabla_{B}  \sum_{i=1}^{O}L(W',H', d^{(\textrm{val})}_i)\\
    & = & B- \xi_{B}  \sum_{i=1}^{O}
    (\frac{\partial W'}{\partial B} \nabla_{W'} L(W',H', d^{(\textrm{val})}_i)+\frac{\partial H'}{\partial B} \nabla_{H'} L(W',H', d^{(\textrm{val})}_i)),\\
    \label{eq:update_b}
    \end{array}
\end{equation}
where 
\begin{equation}
    \frac{\partial W'}{\partial B}= -\xi_{W} \gamma  \nabla^2_{B,W} \sum\limits_{i=1}^{M} b_i L(W,H, d^{(\textrm{pre})}_i),
\end{equation}
and
\begin{equation}
    \frac{\partial H'}{\partial B}= -\xi_{H} \gamma  \nabla^2_{B,H} \sum\limits_{i=1}^{M} b_i L(W,H, d^{(\textrm{pre})}_i).
\end{equation}

\begin{algorithm}[H]
\SetAlgoLined
 \While{not converged}{
1. Update encoder $V$ in  pretraining phrase using Eq.(\ref{eq:update_v})\\
2. Update  head $J$ in  pretraining  using Eq.(\ref{eq:update_j})\\
3. Update encoder $W$ in  finetuning phrase using Eq.(\ref{eq:update_w})\\
4. Update  head $H$ in  finetuning  using Eq.(\ref{eq:update_h})\\
5. Update ignoring variables $A$ using Eq.(\ref{eq:update_a})
 }
 \caption{Optimization algorithm for LBI}
 \label{algo:algo}
\end{algorithm}

\begin{table*}[t]
\caption{Accuracy (\%) on the Office31 dataset. In the $A\to B$ notion, $A$ denotes the source data and $B$ denotes the target data.  1, 2, 3 in DAN, DANN, CDAN, MME denote  unsupervised, semi-supervised, and supervised domain adaptation settings respectively.}
\centering
\begin{adjustbox}{width=\columnwidth,center}
\begin{tabular}{l|ll|llllll|c}
\toprule
\multicolumn{3}{l}{} & A$\to$W	& A$\to$D	&D$\to$W &	D$\to$A &	W$\to$D	& W$\to$A&	Average\\
 \hline
\multicolumn{3}{l|}{DAN-1~\citep{long2015learning}}
&78.11&67.27&91.12&50.66&98.18&53.85&73.20\\
\multicolumn{3}{l|}{DAN-2~\citep{long2015learning}}&95.86&98.18&96.45&86.12&97.27&87.05&93.49\\
\multicolumn{3}{l|}{DAN-3~\citep{long2015learning}}&97.04&96.36&97.04&\textbf{88.74}&\textbf{100.0}&87.24&94.40\\
 \multicolumn{3}{l|}{DANN-1~\citep{ganin2016domain}}&78.70&68.18&87.57&46.72&94.55&50.28&71.00\\
\multicolumn{3}{l|}{DANN-2~\citep{ganin2016domain}}&94.08&93.64&94.67&85.55&95.45&84.99&91.40\\
\multicolumn{3}{l|}{DANN-3~\citep{ganin2016domain}}&96.45&94.55&\textbf{97.63}&86.68&\textbf{100.0}&85.37&93.45\\
\multicolumn{3}{l|}{CDAN-1~\citep{long2017conditional}}&83.84&71.82&94.08&56.47&97.27&60.98&77.41\\
\multicolumn{3}{l|}{CDAN-2~\citep{long2017conditional}} &94.67&95.45&96.45&86.68&97.27&84.80&92.55\\
\multicolumn{3}{l|}{CDAN-3~\citep{long2017conditional}} &97.04&95.45&\textbf{97.63}&85.93&99.09&87.24&93.73\\
\multicolumn{3}{l|}{MME-1~\citep{saito2019semi}}&60.95&50.91&88.17&40.71&95.45&41.28&62.91\\
\multicolumn{3}{l|}{MME-2~\citep{saito2019semi}}&95.86&96.36&96.45&84.05&99.09&85.74&92.93\\
\multicolumn{3}{l|}{MME-3~\citep{saito2019semi}} &94.67&90.90&\textbf{97.63}&87.24&\textbf{100.0}&84.43&92.48\\
\midrule
&Pretrain & Finetune & 
\multicolumn{7}{l}{}\\
\hline
Ablation 1&No  & No source&96.45&97.48&\textbf{97.63}&85.46&99.09&85.37&93.58\\
Ablation 2&No& Full source&96.45&93.64&\textbf{97.63}&87.43&99.09&86.12&93.39\\
Ablation 3&No& Weighted source&98.22&97.27&\textbf{97.63}&86.68&98.18&86.87&94.14\\
Ablation 4&Full source& No source &97.04&97.27&\textbf{97.63}&85.74&98.18&86.38&93.71\\
Ablation 5&Full source& Full source  &95.86&95.45&97.04&87.05&\textbf{100.0}&86.30&93.62\\
Ablation 6&Full source& Weighted source &97.63&96.36&\textbf{97.63}&88.18&99.09&\textbf{87.99}&94.48\\
Ablation 7&Weighted source& No source&96.45&\textbf{99.09}&97.04&86.49&99.09&86.49&94.11\\
Ablation 8&Weighted source& Full source&95.86&91.82&\textbf{97.63}&88.37&\textbf{100.0}&87.05&93.46\\
\hline
Full LBI (ours) &Weighted source &Weighted source  &\textbf{98.82}&\textbf{99.09}&97.04&87.80&99.09&86.87&\textbf{94.79}\\
\bottomrule
\end{tabular}
\end{adjustbox}
\label{tb:results-31}
\end{table*}

\begin{table*}[t]
\caption{Accuracy (\%) on the ImageCLEF dataset. In the $A\to B$ notion, $A$ denotes the source data and $B$ denotes the target data.  1, 2, 3 in DAN, DANN, CDAN, MME denote  unsupervised, semi-supervised, and supervised domain adaptation settings respectively.}
\centering
\begin{adjustbox}{width=0.99\columnwidth,center}
\begin{tabular}{lll|llllll|c}
\toprule
&& &  C$\to$P	&C$\to$I&	P$\to$C&	P$\to$I&	I$\to$C&	I$\to$P&	Average\\
 \hline

\multicolumn{3}{l|}{DAN-1~\citep{long2015learning}}&68.15&76.40&89.19&78.88&88.51&66.88&78.00\\
\multicolumn{3}{l|}{DAN-2~\citep{long2015learning}}&61.78&83.85&93.92&83.85&92.57&63.06&79.84\\
\multicolumn{3}{l|}{DAN-3~\citep{long2015learning}}&64.33&81.37&93.25&85.71&93.24&68.79&81.12\\
 \multicolumn{3}{l|}{DANN-1~\citep{ganin2016domain}}&61.78&73.29&84.46&77.02&75.68&66.24&73.08\\
 \multicolumn{3}{l|}{DANN-2~\citep{ganin2016domain}}&63.06&83.23&93.92&85.71&93.24&64.33&80.58\\
 \multicolumn{3}{l|}{DANN-3~\citep{ganin2016domain}}&68.43&83.85&95.27&88.20&91.89&69.26&82.82\\
\multicolumn{3}{l|}{CDAN-1~\citep{long2017conditional}}&64.97&70.81&70.27&78.26&90.54&64.33&73.20\\
\multicolumn{3}{l|}{CDAN-2~\citep{long2017conditional}} &64.42&85.71&95.27&85.09&95.27&67.52&82.21\\
\multicolumn{3}{l|}{CDAN-3~\citep{long2017conditional}} &64.97&87.58&93.24&\textbf{89.44}&94.59&67.52&82.89\\
\multicolumn{3}{l|}{MME-1~\citep{saito2019semi}}&57.96&68.32&75.00&72.05&85.14&64.97&70.57\\
\multicolumn{3}{l|}{MME-2~\citep{saito2019semi}}&61.15&80.12&93.92&81.99&93.92&59.97&78.51\\
\multicolumn{3}{l|}{MME-3~\citep{saito2019semi}} &64.97&80.75&88.51&84.47&91.22&61.78&78.62\\
\midrule
&Pretrain & Finetune & 
\multicolumn{7}{l}{}\\
\hline
Ablation 1&No  & No source&59.87&86.34&91.89&84.47&91.22&65.61&79.90\\
Ablation 2&No& Full source&68.02&86.34&93.24&86.04&90.54&67.39&81.93\\
Ablation 3&No& Weighted source&67.52&88.20&94.59&84.47&95.95&66.88&82.94\\
Ablation 4&Full source& No source &62.42&87.28&93.87&83.79&93.72&63.43&80.75\\
Ablation 5&Full source& Full source &68.79&86.96&94.59&85.09&93.92&68.15&82.92\\
Ablation 6&Full source& Weighted source&68.15&\textbf{89.44}&\textbf{95.45}&86.95&\textbf{97.30}&68.15&84.24
\\
Ablation 7&Weighted source& No source&63.06&87.58&93.23&86.34&93.24&64.33&81.30\\
Ablation 8&Weighted source& Full source&66.88&85.09&93.24&85.71&94.59&66.88&82.07\\
\hline
Full LBI (ours) &Weighted source &Weighted source &\textbf{70.70}&\textbf{89.44}&93.24&87.58&95.95&\textbf{70.06}&	\textbf{84.50}\\
\bottomrule
\end{tabular}
\end{adjustbox}
\label{tb:results-clef}
\end{table*}

\begin{table*}[t]
\caption{Accuracy (\%) on the OfficeHome dataset. 
}
\centering
\begin{adjustbox}{width=\columnwidth,center}
\begin{tabular}{l|ll|llllll|c}
\toprule
\multicolumn{3}{l|}{}& Ar$\to$Cl & Ar$\to$Pr & Ar$\to$Rw &  Cl$\to$Ar & Cl$\to$Pr & Cl$\to$Rw & Pr$\to$Ar\\
 \hline
\multicolumn{3}{l|}{DAN-1~\citep{long2015learning}}	&30.86&	46.25&	54.19&	33.54&	41.22&45.03&33.33\\
\multicolumn{3}{l|}{DAN-2~\citep{long2015learning}}	&67.66&	86.77&	72.74&	57.51&	87.59&69.50&62.14\\
\multicolumn{3}{l|}{DAN-3~\citep{long2015learning}}	&68.46&	\textbf{87.94}&	75.53&	59.05&	85.13&72.29&60.29\\
 \multicolumn{3}{l|}{DANN-1~\citep{ganin2016domain}} &	33.37&43.56&55.75&	40.95	&47.66	&50.50&37.24\\
 \multicolumn{3}{l|}{DANN-2~\citep{ganin2016domain}} &	67.09&	83.37&	73.63&	55.56&	82.79&69.83&58.64\\
 \multicolumn{3}{l|}{DANN-3~\citep{ganin2016domain}} &	68.80&	85.60&	72.96&	60.08&	86.53&72.74&59.88\\
\multicolumn{3}{l|}{CDAN-1~\citep{long2017conditional}}	&37.49	&49.77&	59.55&	41.56&	51.99&54.41&41.77\\
\multicolumn{3}{l|}{CDAN-2~\citep{long2017conditional}}&	68.23&	84.43&	73.41&	60.49&	84.66&71.84&59.67\\
\multicolumn{3}{l|}{CDAN-3~\citep{long2017conditional}} &	67.31	&86.65&	73.52&	61.93&	84.66&71.06&61.93\\
\multicolumn{3}{l|}{MME-1~\citep{saito2019semi}}	&28.11&	40.40	&52.40	&29.84&	35.71&40.34&29.22\\
\multicolumn{3}{l|}{MME-2~\citep{saito2019semi}}&	68.23	&84.66&	70.73&	55.35&	85.01&66.37&58.44\\
\multicolumn{3}{l|}{MME-3~\citep{saito2019semi}}&	69.60&	85.01&	71.96	&56.97&	84.66&67.93&52.47\\
\hline
&Pretrain & Finetune & 
\multicolumn{7}{l}{}\\
\hline
Ablation 1&No  & No source&66.40&87.47&73.18&65.23&87.24&72.96&64.40\\
Ablation 2&No& Full source &68.69&85.36&74.75&60.08&86.07&72.29&59.47\\
Ablation 3&No& Weighted source&68.00&86.42&75.08&62.14&86.53&73.52&63.37\\
Ablation 4&Full source& No source&69.71&\textbf{87.94}&\textbf{76.65}&67.28&88.17&76.54&65.02\\
Ablation 5&Full source& Full source  &67.54&85.13&74.75&60.08&84.54&71.17&61.11\\
Ablation 6&Full source& Weighted source &68.00&84.66&74.64&61.73&86.3&71.84&63.17\\
Ablation 7&Weighted source& No source&\textbf{70.51}&87.70&76.20&\textbf{69.75}&\textbf{89.34}&\textbf{77.65}&\textbf{68.52}\\
Ablation 8&Weighted source& Full source&67.66&85.25&74.41&61.52&86.77&73.30&61.52\\
\hline
Full LBI (ours) &Weighted source &Weighted source&69.03&84.66&75.42&61.32&85.83&73.52&62.76\\
 \hline
 \hline
\multicolumn{3}{l|}{}&   Pr$\to$Cl & Pr$\to$Rw & Rw$\to$Ar & Rw$\to$Cl & Rw$\to$Pr &\multicolumn{2}{|c}{Average}   \\
 \hline
\multicolumn{3}{l|}{DAN-1~\citep{long2015learning}}&28.69&52.96&51.03&33.37&65.69&\multicolumn{2}{|c}{43.01}   \\
\multicolumn{3}{l|}{DAN-2~\citep{long2015learning}}&70.40&74.19&65.23&68.11&86.07&\multicolumn{2}{|c}{72.33}   \\
\multicolumn{3}{l|}{DAN-3~\citep{long2015learning}}&70.51&75.75&63.79&69.14&86.89&\multicolumn{2}{|c}{72.90}   \\
 \multicolumn{3}{l|}{DANN-1~\citep{ganin2016domain}}&33.37&56.54&52.67&42.74&70.73&\multicolumn{2}{|c}{47.09}   \\
 \multicolumn{3}{l|}{DANN-2~\citep{ganin2016domain}}&65.60&72.29&60.70&68.00&86.07&\multicolumn{2}{|c}{70.30}   \\
 \multicolumn{3}{l|}{DANN-3~\citep{ganin2016domain}}&68.80&73.18&64.61&68.46&87.70&\multicolumn{2}{|c}{72.45}   \\
\multicolumn{3}{l|}{CDAN-1~\citep{long2017conditional}}&41.14&58.21&56.58&41.83&73.54&\multicolumn{2}{|c}{50.65}   \\
\multicolumn{3}{l|}{CDAN-2~\citep{long2017conditional}}&69.49&73.41&63.58&67.54&88.06&\multicolumn{2}{|c}{72.07}   \\
\multicolumn{3}{l|}{CDAN-3~\citep{long2017conditional}} &\textbf{70.74}&72.63&65.23&69.94&87.35&\multicolumn{2}{|c}{72.75}   \\
\multicolumn{3}{l|}{MME-1~\citep{saito2019semi}}&26.97&48.04&46.71&33.71&61.71&\multicolumn{2}{|c}{39.43}   \\
\multicolumn{3}{l|}{MME-2~\citep{saito2019semi}}&67.31&69.50&59.26&66.51&86.18&\multicolumn{2}{|c}{69.80}   \\
\multicolumn{3}{l|}{MME-3~\citep{saito2019semi}}&69.14&73.07&62.96&68.80&85.83&\multicolumn{2}{|c}{70.70}   \\
\hline
&Pretrain & Finetune & 
\multicolumn{7}{l}{}\\
\hline
Ablation 1&No  & No source&68.91&73.85&60.70&69.14&86.77&\multicolumn{2}{|c}{73.02}\\
Ablation 2&No& Full source &67.77&74.75&66.16&69.60&87.47&\multicolumn{2}{|c}{72.71}\\
Ablation 3&No& Weighted source&69.94&74.53&65.02&\textbf{70.17}&89.23&\multicolumn{2}{|c}{73.66}\\
Ablation 4&Full source& No source&69.71&76.20&\textbf{68.96}&69.37&88.06&\multicolumn{2}{|c}{75.30}\\
Ablation 5&Full source& Full source &66.74&71.84&64.61&66.63&87.00&\multicolumn{2}{|c}{71.76}\\
Ablation 6&Full source& Weighted source&66.97&74.19&65.43&67.54&88.41&\multicolumn{2}{|c}{72.74}\\
Ablation 7&Weighted source& No source&70.63&\textbf{77.65}&67.90&69.83&\textbf{90.52}&\multicolumn{2}{|c}{\textbf{76.35}}\\
Ablation 8&Weighted source& Full source&67.43&74.19&68.31&68.34&86.18&\multicolumn{2}{|c}{72.91}\\
\hline
Full LBI (ours) &Weighted source &Weighted source&69.94&74.41&67.28&67.89&86.18&\multicolumn{2}{|c}{73.19}\\
\bottomrule
\end{tabular}
\end{adjustbox}
\label{tb:results-home}
\end{table*}

\section{Experiments}
In this section, we present experimental results. Please refer to the supplements for more  hyperparameter settings and additional results.
\subsection{Datasets}
We perform experiments on three datasets:  OfficeHome~\citep{venkateswara2017deep} Office31~\citep{saenko2010adapting}, and ImageCLEF~\citep{muller2010imageclef}. OfficeHome consists of 15,500 images of daily objects from 65 categories and 4 domains, including Art (Ar), Clipart (Cl), Product (Pr), and Real-World (Rw). 
Each category has an average of about 70 images and  a maximum of 99 images. Office31 contains 4,110 images belonging to 31 classes and 3 domains, including Amazon website (A), web camera (W), and digital SLR camera (D). The number of images in domain A, W, and D is 2817, 795, and 498 respectively. ImageCLEF consists of 1,800 images from 3 datasets (domains) with 12 classes, including Caltech-256 (C), ILSVRC2012 (I), and Pascal VOC2012 (P).  We split OfficeHome, Office31, and ImageCLEF into train/validation/test sets with a ratio of 5:3:2, 6:2:2, and 6:2:2 respectively. For each dataset, we perform domain adaptation~\citep{long2015learning} studies: one domain is selected as source and another domain is selected as target; data in the source domain is leveraged to help with model training in the target domain. In the learning-by-ignoring framework, source data is used as pretraining data and target data is used as finetuning data. 

\subsection{Baselines}
We compare with the following baselines. For each baseline, we experimented with three domain adaptation (DA) settings: 1) unsupervised DA, where the labels of source examples are used for DA and the labels of target examples are not; 2) semi-supervised DA, where the  labels of target examples are used for DA and the labels of source examples are not; 3) supervised DA, where both the labels of source examples and target examples are used for DA. 
\begin{itemize}
\setlength\itemsep{0em}
\item \textbf{DAN}~\citep{long2015learning}, which matches mean embeddings of different domain distributions in a reproducing kernel Hilbert space.
\item \textbf{DANN}~\citep{ganin2016domain}, which uses adversarial learning to learn representations so that source domain and target domain are not distinguishable.
\item \textbf{CDAN}~\citep{long2017conditional}, which conditions  adversarial adaptation  on discriminative information.
    \item \textbf{MME}~\citep{saito2019semi}, which is a semi-supervised domain adaptation method based on minimax entropy.
\end{itemize}

\subsection{Experimental Settings}
We use ResNet-34~\citep{resnet} pretrained on ImageNet~\citep{deng2009imagenet} as the backbone of our methods. Our models were
trained using the Adam~\citep{adam} optimizer, with a batch size of 64, a learning rate of 1e-4 for the feature extractor, a learning rate of 1e-3 for the classifier, a weight decay of 5e-4, for 50 epochs. The learning rate was decreased by a factor of 10 after 40 epochs. In learning by ignoring, $\gamma$ is set to 1 for all datasets; $\lambda$ is set to 7e-3 for OfficeHome, 5e-4 for Office31, and 3e-3 for ImageCLEF.

\subsection{Results}
Table~\ref{tb:results-31} shows the results on the Office31 dataset. In the $A\to B$ notion, $A$ denotes the source dataset and $B$ denotes the target dataset. As can be seen, our proposed LBI method achieves better averaged accuracy than the domain adaptation (DA) baselines including DAN, DANN, CDAN, and MME. The major reason is that our method automatically identifies source examples that are not suitable for pretraining and excludes them from the pretraining process. In contrast, these DA baselines adapt all source examples into the target domain and lack the ability of explicitly identifying source examples that are not suitable for domain adaptation. Table~\ref{tb:results-31} shows the results on the ImageCLEF dataset. Our proposed LBI achieves better average accuracy than the DA baselines, thanks to its mechanism of ignoring source examples that are not suitable for helping with the learning on the target domain. Table~\ref{tb:results-home} shows the results on the OfficeHome dataset. LBI outperforms all DA methods, which further demonstrates the effectiveness of our proposed learning-by-ignoring framework. On this dataset, using source dataset for finetuning leads to worse performance. This is probably because the source domains in this dataset have large domain shifts with the target domains. Therefore, using source data directly for finetuning may not be proper. However, using non-ignored source data for pretraining is helpful.

\subsection{Ablation Studies}
We perform ablation studies to check the effectiveness of individual modules in our framework. Unless otherwise notified, $\gamma$ was set to 1 for all three datasets; $\lambda$ was set to 7e-3 for OfficeHome, 5e-4 for Office31, and 3e-3 for ImageCLEF. 
\begin{itemize}
\setlength\itemsep{0em}
\item Ablation setting 1: no pretraining, finetune on target data only. We ignore the source dataset and directly train a model on the target dataset. This is equivalent to setting both $\lambda$ and $\gamma$ in LBI  (Eq.(\ref{eq:lbi-2})) to 0. 
    \item Ablation setting 2: no pretraining, finetune on target data and all source examples.
    There is no pretraining; we combine the source dataset and the target dataset as a single dataset, then train a model on the combined dataset. This is equivalent to setting $\lambda$ to 0 and setting all ignoring variables in $B$ to 1 in  LBI  (Eq.(\ref{eq:lbi-2})). 
\item Ablation setting 3: no pretraining, finetune on target data and weighted source examples. There is no pretraining; the model is  directly trained on all  target examples and selected source examples. This is equivalent to setting the tradeoff parameter $\lambda$ to 0 in the LBI framework (Eq.(\ref{eq:lbi-2})). 
   \item Ablation setting 4: pretrain on all source examples, finetune on target data only. 
   We first train a model $M_1$ on the full source dataset. Then we train a model $M_2$ on the target dataset. When training $M_2$, we encourage its weights to be close to the optimally trained weights of $M_1$ by minimizing their squared L2 distance. This is equivalent to setting $\gamma$ to 0 and setting all ignoring variables in $A$ to 1 in  LBI  (Eq.(\ref{eq:lbi-2})). 
   \item Ablation setting 5: pretrain on all source examples, finetune on target data and all source examples. This setting is similar to ablation 4, except that $M_2$ is trained on the target and full source dataset. 
   This is equivalent to  setting all ignoring variables in $A$ and $B$ to 1 in  LBI  (Eq.(\ref{eq:lbi-2})).

\item Ablation setting 6: pretrain on all source examples, finetune on target data and weighted source examples. This setting is similar to ablation 4, except that $M_2$ is trained on the target and weighted source dataset. 
This is equivalent to  setting all ignoring variables in $A$  to 1 in  LBI  (Eq.(\ref{eq:lbi-2})). 
\item Ablation setting 7: pretrain on weighted source examples, finetune on target data only. 
 This is equivalent to setting the tradeoff parameter $\gamma$ to 0 in the LBI framework (Eq.(\ref{eq:lbi-2})). 
\item Ablation setting 8: pretrain on weighted source examples, finetune on target data and all source examples. In this setting, all source examples are used for finetuning, without ignoring any of them. This is equivalent to setting all ignoring variables in $B$ to 1 in the LBI framework (Eq.(\ref{eq:lbi-2})).
     \item Ablation study on $\lambda$. We explore how the  performance varies as the tradeoff parameter $\lambda$ in Eq.(\ref{eq:lbi-2}) increases. In this study, the other tradeoff parameter $\gamma$ in Eq.(\ref{eq:lbi-2}) is set to 1. We report the average accuracy on the validation set. The experiments are conducted on ImageCLEF and OfficeHome. Please refer to the supplements for results on OfficeHome.
    \item Ablation study on $\gamma$. We explore how the  performance varies as $\gamma$ increases. We report the average accuracy on the validation set. The experiments are conducted on ImageCLEF and OfficeHome. 
    In this study, the other tradeoff parameter $\lambda$ is set to 3e-3 for ImageCLEF and 7e-3 for OfficeHome. Please refer to the supplements for results on OfficeHome.
\end{itemize}

Table~\ref{tb:results-31} shows the ablation study results on the Office31 dataset. From this table, we make the following observations. \textbf{First}, during pretraining, ignoring certain source examples  is better than using all source examples. For instance, ablation 7 achieves better average accuracy than ablation 4 where the former performs pretraining on weighted source data and the latter uses all source examples for pretraining. The rest settings of ablation 7 and 4 are the same. 
As another example, our proposed full LBI framework achieves higher average accuracy than ablation 6 where the former performs ignoring of certain source examples during pretraining while the latter does not. The rest settings of full LBI are the same as those of ablation 6. The reason is that due to domain shift between source and target, some source examples are largely different from the target examples; if using such source examples for pretraining, the pretrained encoder may be biased to the source distribution and cannot well represent target examples. Our proposed approaches automatically identify source examples that have large domain discrepancy with the target and remove them from the pretraining process. Consequently, the pretrained encoder with ignoring is better than the one pretrained using all source examples without ignoring. \textbf{Second}, when using source examples to finetune the encoder (together with target examples), it is beneficial to ignore some source examples in the finetuning process. As can be seen, in terms of average accuracy, the full LBI framework performs better than ablation 8 where the former ignores certain source examples during finetuning while the latter does not. The rest settings of these two methods are the same. Similarly, ablation 6 outperforms ablation 5; ablation 3 achieves higher average accuracy than ablation 2. Again, the reason is that due to domain shift, some source examples are not suitable for finetuning the encoder of the target dataset. Our proposed ignoring approach excludes such source examples from finetuning, hence achieving better performance than not using ignoring. \textbf{Third}, without the ignoring mechanism, using source examples for finetuning is harmful. For example, ablation 2 which uses all source samples for finetuning achieves lower average accuracy than ablation 1 which uses no source example for finetuning. Similarly, ablation 5 achieves worse average accuracy than ablation 4; ablation 8 performs less well than ablation 7. However, once we add the ignoring mechanism and uses non-ignored source examples for finetuning, the resulting average accuracy is higher than not using any source examples for finetuning, as can be seen from the average accuracy results that ablation 3 outperforms ablation 1; ablation 6 outperform ablation 4; and the full LBI method outperforms ablation 7. This further demonstrates the effectiveness of our proposed ignoring mechanism, which can 1) identify sources examples that are close to the target domain and use them for finetuning; 2) recognize source example having large domain shifts from the target and exclude them from finetuning. In contrast, the baseline methods can only do the black-or-white choice: either using all source examples for finetuning or not using any source example at all, which hence leads to inferior performance.

Table~\ref{tb:results-clef} shows the ablation results on ImageCLEF. From this table, we make similar observations as those in Table~\ref{tb:results-31}. \textbf{First}, in pretraining, the ignoring mechanism which identifies source examples that have large domain shifts from the target and excludes them from the pretraining process achieves better performance than not using ignoring. This is evidenced from the average accuracy results that ablation 7 outperforms ablation 4; our full LBI framework achieves better accuracy than ablation 6. \textbf{Second}, during finetuning, ignoring certain source examples works better than using all source examples. This is demonstrated by the average accuracy results that the full LBI outperforms ablation 8; ablation 6 outperforms ablation 5; and ablation 3 outperforms ablation 2. 

Table~\ref{tb:results-home} shows the ablation results on OfficeHome. The observations from these results are similar to those made in Table~\ref{tb:results-31} and~\ref{tb:results-clef}. \textbf{First}, performing ignoring on source during pretraining is beneficial, as seen from the average accuracy results that ablation 7 outperforms ablation 4; ablation 8 outperforms ablation 5; and the full LBI outperforms ablation 6. \textbf{Second}, performing ignoring on source during finetuning is advantageous than not using ignoring, as demonstrated by the average accuracy results that ablation 3 performs better than ablation 2; ablation 6 outperforms ablation 5; and the full LBI framework outperforms ablation 8. 
\begin{figure}[t]
    \centering
 \includegraphics[width=0.49\columnwidth]{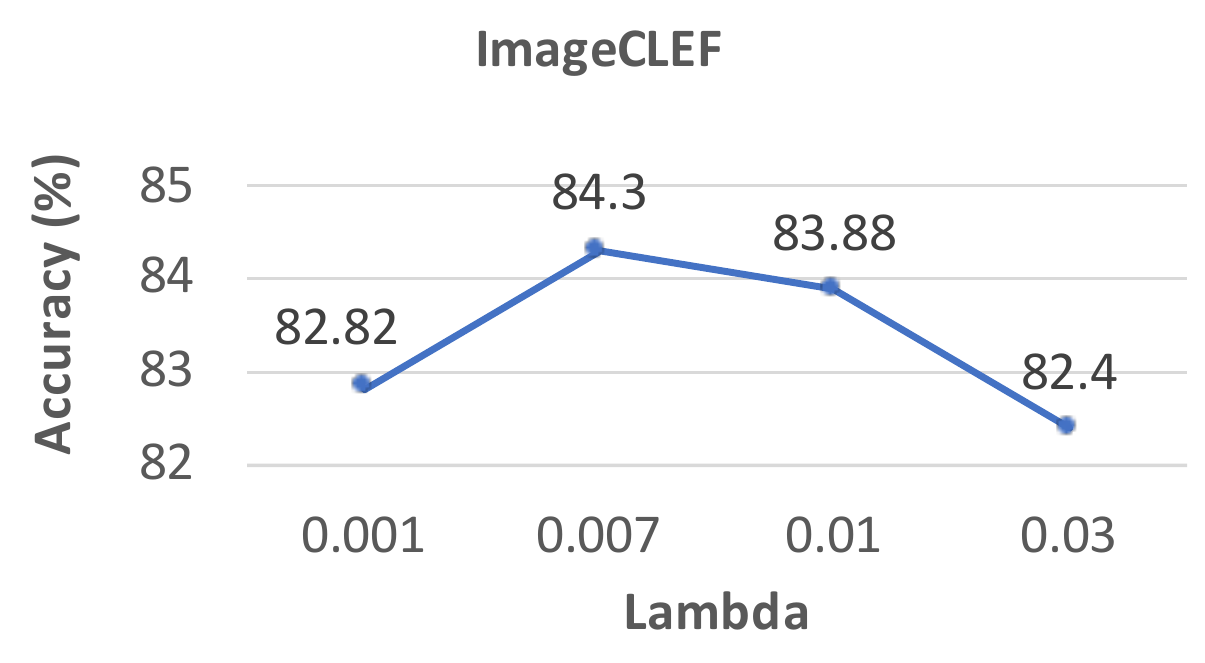}
  \includegraphics[width=0.49\columnwidth]{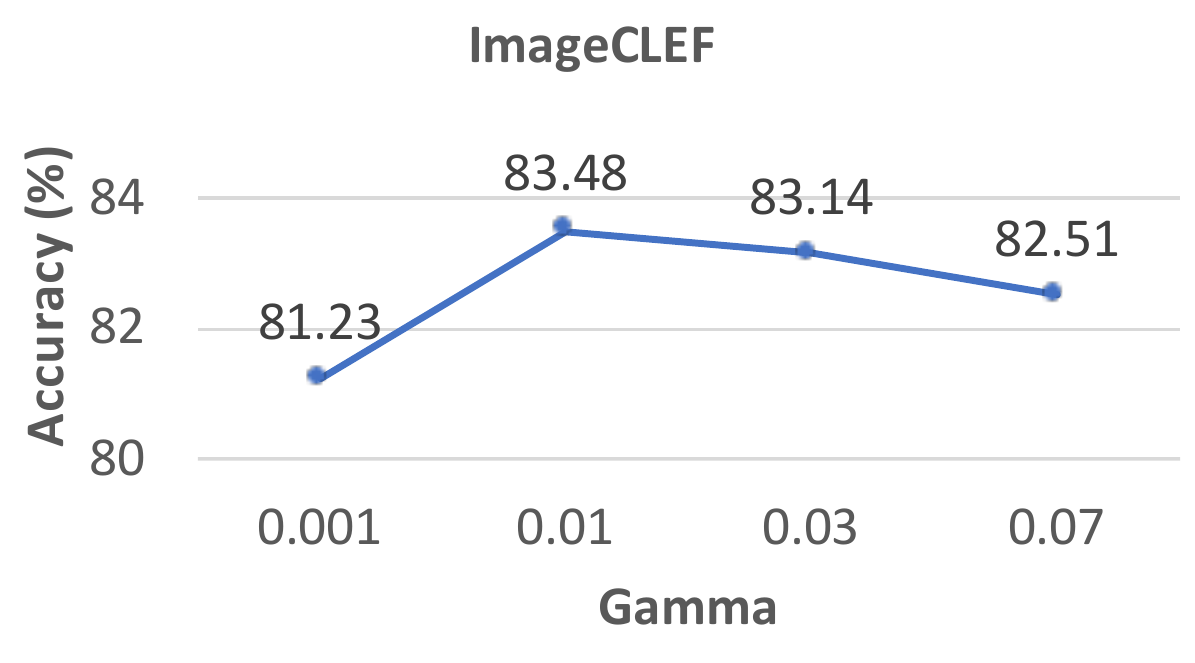}
       \caption{(Left) How accuracy changes as $\lambda$   increases. (Right) How accuracy changes as $\gamma$   increases.}
 \label{fig:lambda}
\end{figure}

Figure~\ref{fig:lambda}(Left) shows how the average accuracy varies as the tradeoff parameter $\lambda$ increases. As can be seen, when $\lambda$ increases from 0.001 to 0.007, the  accuracy increases. This is because a larger $\lambda$ results in a stronger effect of pretraining. The pretrained encoder captures information of non-ignored source examples and such information is valuable for finetuning. However, if $\lambda$ continues to increase, the accuracy starts to decrease. The reason is that an excessively large $\lambda$ leads to too much emphasis on the pretrained encoder and pays less attention to the finetuning on the target data. Figure~\ref{fig:lambda}(Right) shows how the average accuracy varies as the tradeoff parameter $\gamma$ increases. When $\gamma$ increases from 0.001 to 0.01, the accuracy increases. This is because a larger $\gamma$ imposes a stronger utilization of non-ignored source examples as additional data for finetuning. However, further increasing $\gamma$ leads to a worse accuracy. The reason is that if $\gamma$ is too large, the source data will dominate target data. 

\section{Conclusions}
In this paper, we propose a novel machine learning framework -- learning by ignoring (LBI), motivated by the ignoring-driven learning methodology of humans. In LBI, an ignoring variable is learned for each pretraining data example to identify examples that have significant domain difference with the target distribution. 
 We formulate LBI as a three-level optimization problem which consists of three learning stages: an encoder is trained by minimizing a weighted pretraining loss where the loss of each data example is weighted by its ignoring variable; another encoder is finetuned where during the finetuning  this encoder is encouraged to be close to the previously pretrained encoder; the ignoring variables are updated by minimizing the validation loss calculated using the finetuned encoder. We conduct experiments on various datasets where the results demonstrate the effectiveness of our method.

\bibliography{release}

\end{document}